\documentclass[sigconf]{acmart}
\usepackage{amsmath}
\usepackage{amssymb}
\usepackage{amsthm}
\usepackage{xcolor}
\usepackage{graphicx}

\begin{document}

\title{Composing neural algorithms with Fugu}

\author{James B. Aimone*, William Severa, Craig M. Vineyard}
\orcid{0000-0002-7361-253X}
\affiliation{%
  \institution{Center for Computing Research, Sandia National Laboratories}
  \streetaddress{P.O. Box 5800}
  \city{Albuquerque} 
  \state{New Mexico} 
  \postcode{87185-1327}
}
\email{*jbaimon@sandia.gov}

\acmConference[]{}{ }{ } 
\setcopyright{none}
\begin{abstract}
Neuromorphic hardware architectures represent a growing family of potential post-Moore's Law Era platforms.  Largely due to event-driving processing inspired by the human brain, these computer platforms can offer significant energy benefits compared to traditional von Neumann processors.  Unfortunately there still remains considerable difficulty in successfully programming, configuring and deploying neuromorphic systems.  We present the Fugu framework as an answer to this need. Rather than necessitating a developer attain intricate knowledge of how to program and exploit spiking neural dynamics to utilize the potential benefits of neuromorphic computing, Fugu is designed to provide a higher level abstraction as a hardware-independent mechanism for linking a variety of scalable spiking neural algorithms from a variety of sources. Individual kernels linked together provide sophisticated processing through compositionality. Fugu is intended to be suitable for a wide-range of neuromorphic applications, including machine learning, scientific computing, and more brain-inspired neural algorithms.  Ultimately, we hope the community adopts this and other open standardization attempts allowing for free exchange and easy implementations of the ever-growing list of spiking neural algorithms.
\end{abstract}

\maketitle
\section{Introduction}

The success of neuromorphic computing technologies is dependent on its large-scale adoption as a post-Moore's law, low power solution for multiple applications.  The generality of neural computing approaches is an area of active exploration \cite{aimone2019neural, aimone2018non}, but there is already a growing disparity between this value of neuromorphic systems as a general tool and emerging software stacks for leveraging these platforms for specific functions. For individual applications, such as spiking deep neural networks (DNNs), pre-defined specialized solutions are often sufficient. An example of this is the Whetstone software, which we recently introduced \cite{Severa2019Training} as a tool to convert standard Keras-trained DNNs to a spiking-compatible structure.  Similarly, there are an increasing number of options for neural programming environments that are premised on established classes of neural computation.  Tools such as PyNN \cite{davison2009pynn}, Nengo \cite{bekolay2014nengo}, and N2A \cite{rothganger2014n2a} assume certain intent among user communities, and thus while often powerful, these implicitly require that users orient themselves to a certain perspective of neural computation.  

Here, we sought to develop a programming platform to enable the development of neuromorphic applications without substantial knowledge of neural computing or neuromorphic hardware.  Our solution, which we refer to as \textit{Fugu}\footnote{The name \textit{Fugu} is inspired by the Japanese word for pufferfish; which, of course, have spikes. Furthermore, Fugu is considered a culinary delicacy due to the presence of low levels of the neurotoxin tetrodotoxin, or TTX, which has significant value in studying the electrophysiology mechanisms underlying biological action potentials. Only one author (CMV) has eaten Fugu before.}, is intended to facilitate neuromorphic application development in a manner similar to how CUDA facilitates the programming of GPUs.  

Fugu is structured so as to separate the task of programming applications that may leverage neuromorphic hardware from the design of spiking neural algorithms (SNAs) and the specific details of neuromorphic hardware platforms.  We accordingly foresee three categories of users.  The primary target population is the general computer programming community; well-versed in scientific computing approaches but perhaps only modestly familiar with parallel computing and likely unfamiliar with neural approaches.  For the adoption of neuromorphic hardware it is critical that these users can leverage this technology. To enable this, a second group of users - those capable of designing SNAs - also need to be able to incorporate their algorithms into the Fugu construct.  This population of users may be well versed in neural computation generally, but also may not be familiar with specific considerations of different neural hardware platforms.  Finally, the third category of users would be those who are deeply familiar with neuromorphic hardware, and are capable of optimizing and tailoring generic SNAs into algorithmic implementations that are optimized to the current conditions of neuromorphic hardware.  

\section{Background}

Fugu is a high-level programming framework specifically designed for develolping \textit{spiking }algorithms in terms of computation graphs.  At the lowest level, SNAs are directed graphs, with nodes corresponding to neurons and edges corresponding to synapses.  However, by considering how SNAs can be networked together (not unlike how the brain consists of networks of local circuits), Fugu is able to consider a higher level graphical perspective of the overarching computation.

A key underpinning of Fugu is that SNAs, if properly constructed, should be \textit{composable}. In the context here, this means that two appropriately constructed SNAs can be combined to form larger algorithms in that the outputs of one can be used as the inputs to the other.  The following sections will highlight how this concept of compositionality is more than simply matching sizes.  For two small SNAs to be combined into a larger SNA, the sizes, temporal structure, and encoding schemes have to align. This process can become non-trivial when a number of SNAs are to be composed together, the automation of which is the primary contribution of Fugu. 

Fugu is designed primarily for spiking algorithms, the core model for which is described in section \ref{spiking model} below.  However, this emphasis on spiking is not meant to preclude its future use for architectures which may achieve benefits from other features such as analog arrays. Most neural algorithms, whether spiking or not, can be considered through a graphical perspective and thus may be suitable for the approach described here.  

\subsection{Generic spiking neural algorithm model}
\label{spiking model}
Fugu assumes a fairly simple neural model for SNAs, so as to enable the generic use of spiking neuromorphic hardware.  The default neuron model of Fugu is that of leaky-integrate and fire (LIF) neurons, with parameterizable voltage thresholds ($V_{\rm{thresh}}$), time constants $\tau_{i}$, and reset voltages $V_{\rm{reset}}$.  The default synapses are considered point synapses, with a parameterized weight $w_{i,j}$, where $i$ and $j$ are source and target neurons respectively.  At each timestep, each neuron computes the voltage at time $t$ given by $V_t$ in the following manner
\begin{align*}
\hat{V}_j(t) &= \sum_{i} f_{i}(t)*w_{i,j} + V_{j}(t-1),\\
V_j &=  \begin{cases} V_{\rm{reset}} & \text{ if } \hat{V}_j > V_{\rm{thresh}}
\\ (1-\tau_j)\hat{V}_j &\text{ elsewise}. 
\end{cases},\\
f_j(t) &= \begin{cases} P & \text{ if } \hat{V}_j > V_{\rm{thresh}}
\\ 0 &\text{ elsewise}.  \end{cases}
\end{align*}

where the firing, $f_{j}(t)$ is determined if the neuron's voltage crosses a threshold.  To account for probabilistic neural transmission, $P$ is a probabilistic Bernoulli draw of either $0$ or $1$ according to a stochastic firing rate at rate $p$.  If neurons are not stochastic, $p=1$.

\subsection{Learning, structural plasticity, and other extensions to core spiking model}

The design of Fugu allows for capabilities beyond this basic model.  For instance, if a given algorithm requires a learning mechanism with a particular learning coefficient, that can be included as an attribute of specific synapses, but not all neuromorphic hardware may be capable of implementing that algorithm correctly. Similarly, more sophisticated neuron models, such as multi-compartment models (which leverage dendritic dynamics) and conductance-based models are entirely valid from an algorithm design perspective, however there are few hardware platforms that can fully leverage these. 

These added complexities will likely arise with more sophisticated algorithms, particularly from biologically-inspired models.  Fugu thus remains somewhat agnostic to what occurs beyond the basic LIF model within a given SNA, although such functions may present a risk that the algorithm may not be compatible with a downstream platform. Accordingly, one key requirement is that the communication and connections \textit{between }algorithms - which is the operational domain of Fugu, is expressly compatible with the standard LIF model.  In this sense, component SNAs must present discrete spike events as outputs and similarly (with some exceptions) take in spike events as inputs.
 
\section{Design of Fugu}

\begin{figure*}
\includegraphics[width=\textwidth]{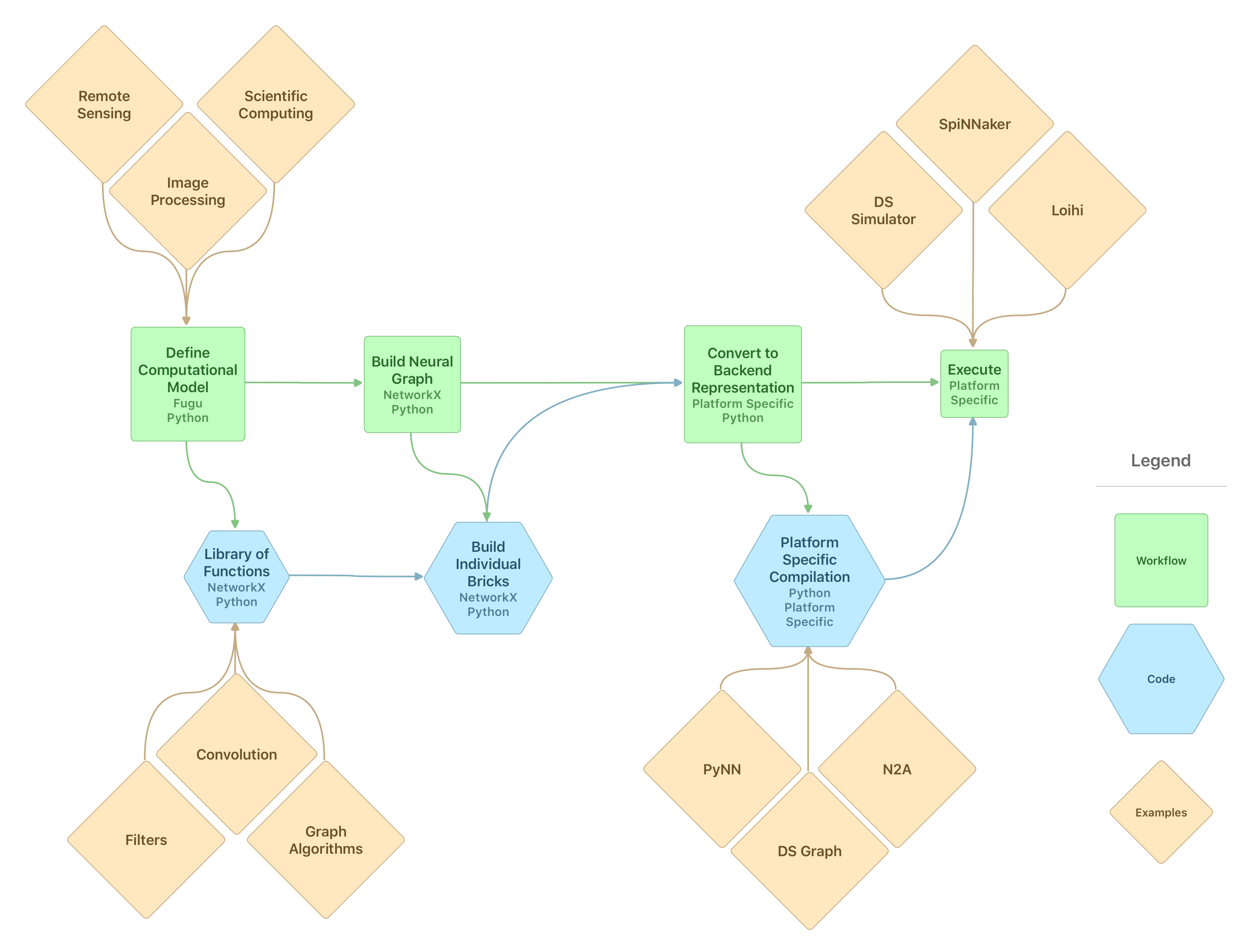}
\caption{An overview diagram of a common workflow in Fugu (Green squares).  Code exists to underlie this workflow (Blue hexagons) with examples provides (Tan diamonds).}
\label{FuguDiagram}
\end{figure*}

As stated above, a key goal of Fugu is to provide a general scientific computing user access to emerging neuromorphic hardware --- specifically spiking neuromorphic hardware --- by providing an accessible library of functions that Fugu can map into neural hardware. Figure \ref{FuguDiagram} provides an overview of the Fugu framework which we will elaborate upon the subcomponents in the following sections. 

Fugu accomplishes this by providing the following core capabilities:
\begin{itemize}
\item \textbf{An API to conventional programming environments (i.e., Python, C++)} 
\item \textbf{Automated construction of a graphical \textit{intermediate representation} of spiking neural algorithms}
\item \textbf{Outputs to neural hardware compilers or Fugu's reference simulator}
\end{itemize}

The following sections expand on each of these topics.

\subsection{API}

The goal of the API for Fugu is to make the construction of a Fugu algorithm be readily called from C++ or Python.  In theory, all of the SNA requirements and processing should be transparent to a user; such that they only have to call a function with standard I/O protocols; not unlike a hardware interface language such as CUDA.

The intent of Fugu is to enable the user to program an application to neuromorphic hardware with only limited knowledge of the underlying architecture or requisite algorithms. Thus, the API level, which is the interface that we expect to be most commonly used, simply requires the user to define a computational graph, which we call a \textit{scaffold}.  Each Fugu scaffold consists of nodes, known as \textit{bricks}, which are the component SNAs, and edges between those bricks that define the flow of information.

\begin{figure}
	\includegraphics[width=2.8in]{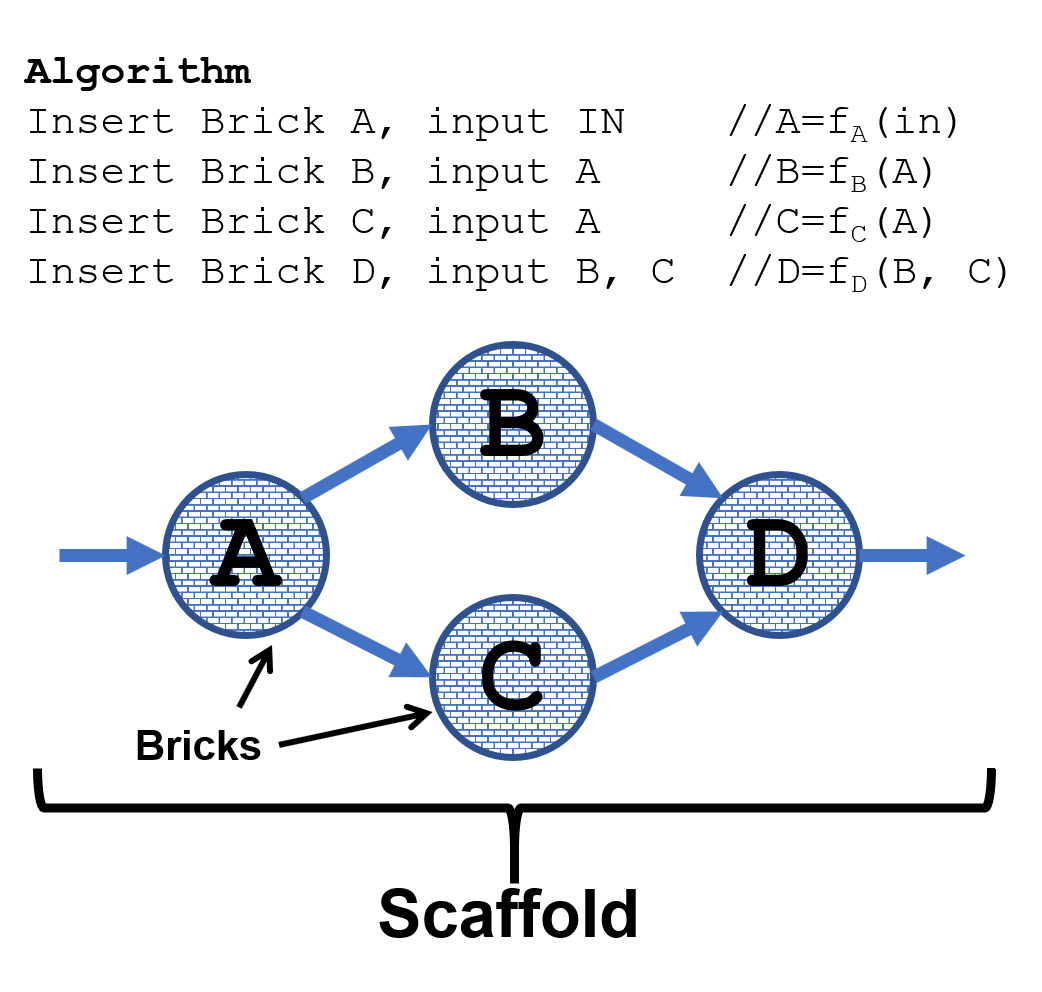}	
	\caption{Fugu Scaffold and Bricks}
	\label{layout}
\end{figure}

During the design of a Fugu application, a user would construct a computational graph of an application.  Consider a simple application with four operations: function \textbf{A} processes an input, functions \textbf{B} and \textbf{C} each process the output of \textbf{A}, and function \textbf{D} combines the outputs of \textbf{B }and \textbf{C}.  As shown in the pseudocode within Figure \ref{layout}, from a simple set of instructions of how to connect these four functions, Fugu would construct bricks for each function and compose them into a larger algorithm scaffold.  

This scaffold is a graphical representation of the desired Fugu algorithm, however this perspective remains agnostic the eventual populations of neurons that will perform the desired computation.  Each of the bricks consists of instructions for building its particular neural algorithm, which can take any number of forms.  To become functional, the overall scaffold must progress from individual brick representations which must be populated with the appropriate internal neural circuits that have been appropriately scaled and configured to interface with all of the brick's neighbors.


\subsection{Intermediate Representation}

The primary contribution of Fugu is the managed and automated IR between higher-level coding environments and low-level neuromorphic hardware and their compilers.   This IR consists of three components that, during compilation, provide the connection between the API and the compiler output: a library of SNA \textit{bricks}, a collection of algorithms for linking SNA bricks, and the combined application graph output.  

The IR of Fugu exists within Python, and it leverages the NetworkX libary \cite{hagberg2008exploring} to construct and manage the neural circuits that will be generated and combined during Fugu operation.

\subsubsection{Library of SNA Bricks}
The Fugu library consists of a growing set of SNA bricks that are suitable for being composed together into larger functional units.  Importantly, the algorithm bricks that are contained within Fugu generally do not consist of explicit neural circuits, but rather they are scripts that can generate the appropriate neural circuit for a given application.  This allows them to be sized appropriately to tailor them to interface with predecessors.

\begin{figure}
	\includegraphics[width=2.8in]{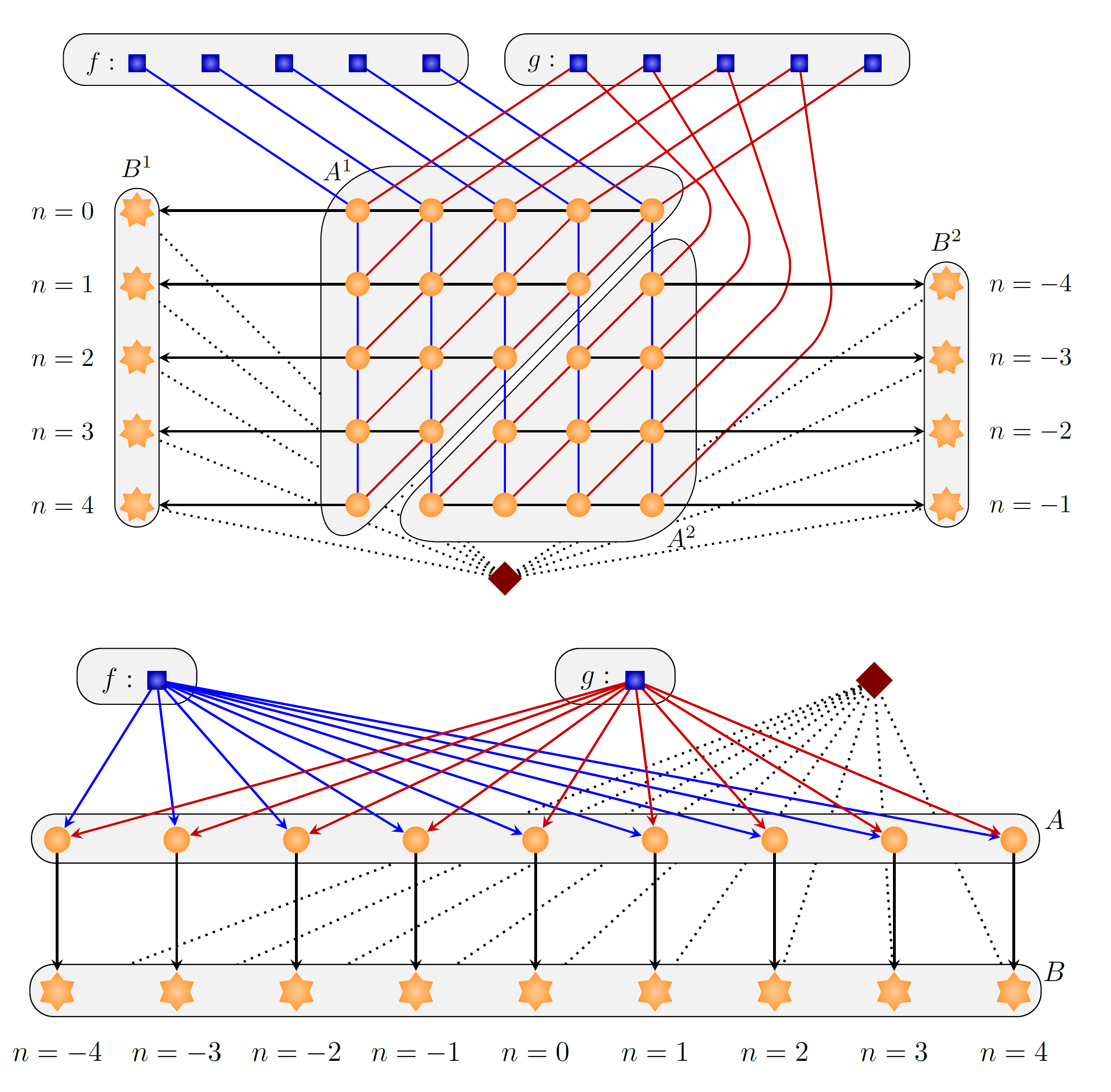}
	\caption{Two SNAs for computing cross-correlation requring different space and time complexities}
	\label{cross_correlation}
\end{figure}

For example, we consider the constant-time 1-dimension max cross-correlation algorithm in \cite{severa2016spiking}.  That algorithm compares two binary vectors of length $N$ by having a dedicated neuron within an intermediate layer calculate each potential off-set, requiring an intermediate layer of size $N^2$.  Subsequently, an output layer, sized $2N-1$ samples that intermediate layer to determine the relative shift between inputs that yields the maximal overlap. Figure \ref{cross_correlation} illustrates these two different SNAs with their differing resource requirements.   

As the description shows, the exact neural circuit needed is highly specific to $N$.  In addition, the resulting circuits have other important properties that will be necessary at later Fugu steps.  For instance, for the constant-time algorithm, the inputs are provided all at the same time ($T_{in} = 1$), and the output is always a single time-step ($T_{out} = 1$) arriving two time steps later ($D = 2$).  An alternative version of this algorithm streams the inputs in over time and uses delays to make the same computation with fewer neurons, albeit at an extended time cost, a characteristic that will produce different neural circuit and associated metadata.  It is also important to note that this metadata may be a function of input parameters as well --- for instance in the matrix-multiplication application described in \cite{parekh2018constant}, there is a version of the algorithm whose depth is $O(\log \log N)$, where $N$ is the size of the largest matrix dimension.  

\begin{figure}
	\includegraphics[width=2.8in]{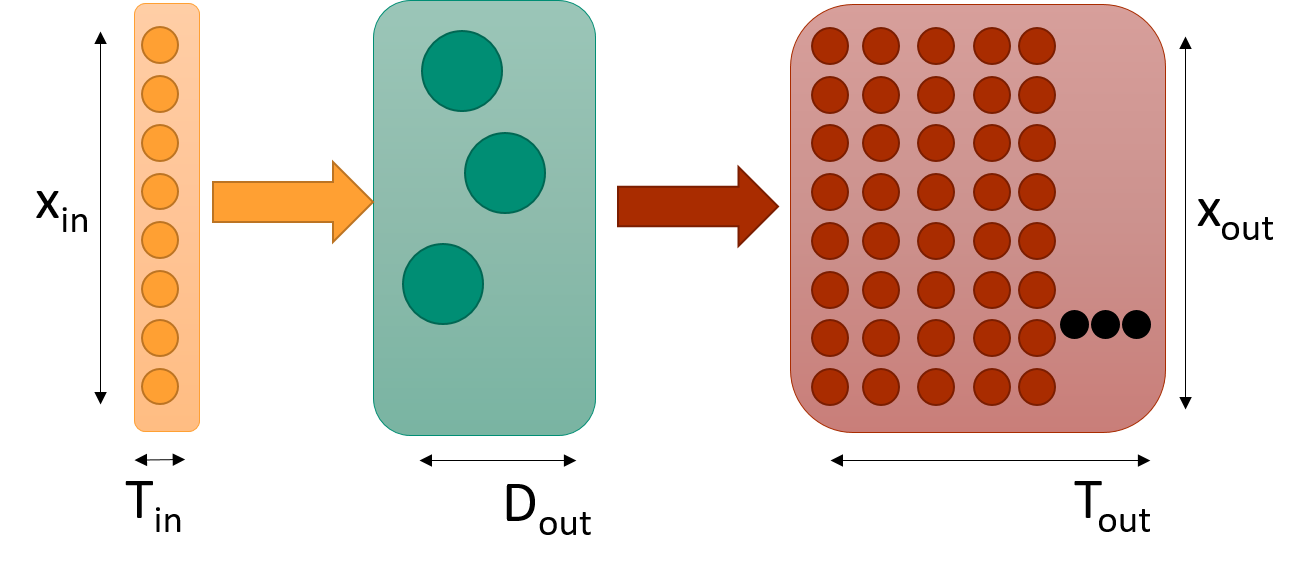}
	\caption{Spiking Algorithm as Fugu Brick}
	\label{fig_brick}
\end{figure}

A schematic of a Fugu brick is shown in Figure \ref{fig_brick}. Some properties common for bricks within Fugu are:

\begin{itemize}
\item $N_{in}$ -- number of input neurons
\item $T_{in}$ -- time length of inputs (how long do inputs stream in).  =inf if streaming
\item $N_{out}$ -- number of output neurons
\item $T_{out}$ -- time length of output.  = inf if streaming
\item $D$ -- circuit depth, corresponding to how many global timesteps must pass for the input at t=1 to reach  
\end{itemize}

However, ultimately the determination of the required parameters is dependent on the coding scheme.  To abstract these concepts, we incorporate a local `index' to each of a brick's output neurons.  A neuron's index indicates (to any downstream brick) the information represented by that neuron (e.g., which component of a coding is represented by that neuron).  By adopting a standardization on this indexing, bricks are able to communicate with one another without imposing design considerations within the bricks themselves.  Additionally, the index provides a flexible mechanism supporting an arbitrary number of dimensions or subsets.

To exist within Fugu, each of the bricks must be able to take instructions from the Fugu environment and produce the appropriate neural circuit NetworkX graph per the necessary scales and timings.  These circuits are referred to as \textit{local circuits}, and are standalone circuit SNAs for computing the requisite function.


\subsubsection{Linking code to combine local circuits into global circuit}

Once a scaffold is constructed by linking bricks together, the scaffold must build a comprehensive global graph (which we call a \textit{circuit}).  This circuit is a platform-independent intermediate representation that becomes complied down to platform-specific code.  Fugu builds the circuit using a lazy method, iterating over the nodes of the scaffold graph.  When a brick is capable of building itself (due to the determination of parameters upstream), it builds its local graph according to the build-time requirements of the scaffold and this local graph is incorporated into the global circuit.  The process of laying bricks is seamless and automatic to the end user, and brick developers only need to manage their own local graphs.  

There are two primary challenges and a number of routine steps that require care for linking these circuits together.

\paragraph{Align sizes of bricks} Each model has an input size $N_{in}$ and an output size $N_{out}$, and in order for two bricks to be compatible with one another \textit{serially}, then it is necessary that the downstream module is scaled appropriately to generate a graph suitably sized to take the outputs.  n general, the shape of the input determines the remainder of the graph.  So, when a user defines the input shape (via input bricks), Fugu can determine the required shape, dimensionality, or other parameters of any connected bricks automatically.  

\begin{figure}
	\includegraphics[width=2.7in]{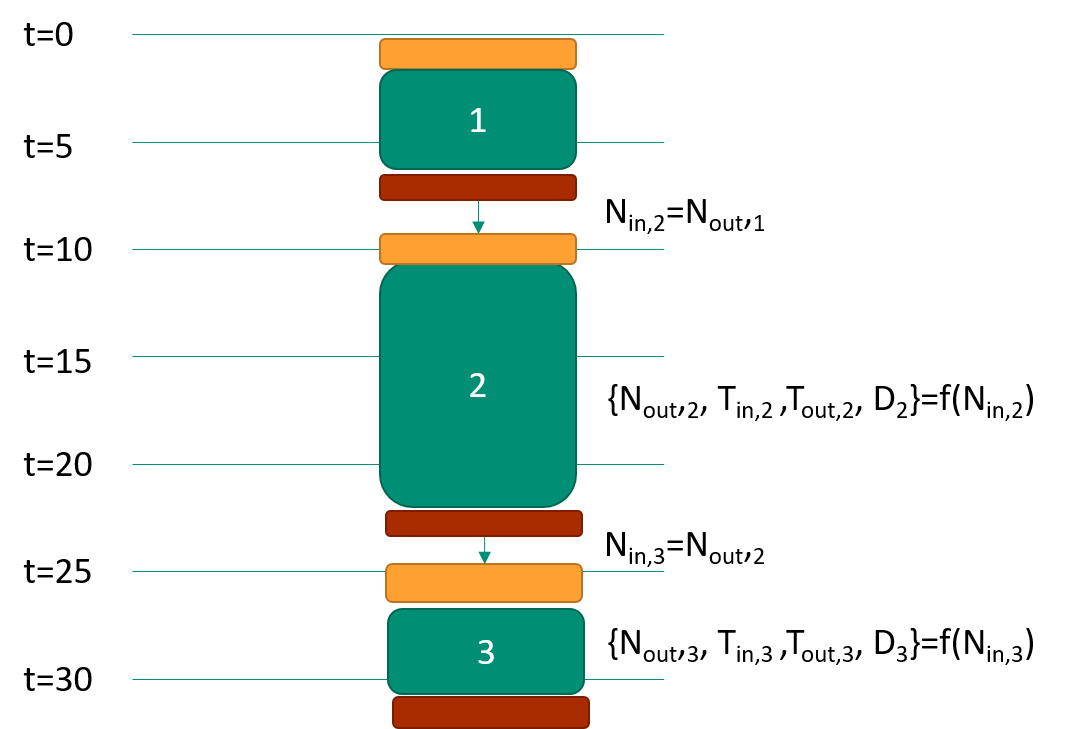}	
	\caption{Normalizing brick sizes}
	\label{serial_link}
\end{figure}

A general scheme is shown in Figure \ref{serial_link}. 

\begin{figure}
	\includegraphics[width=2.8in]{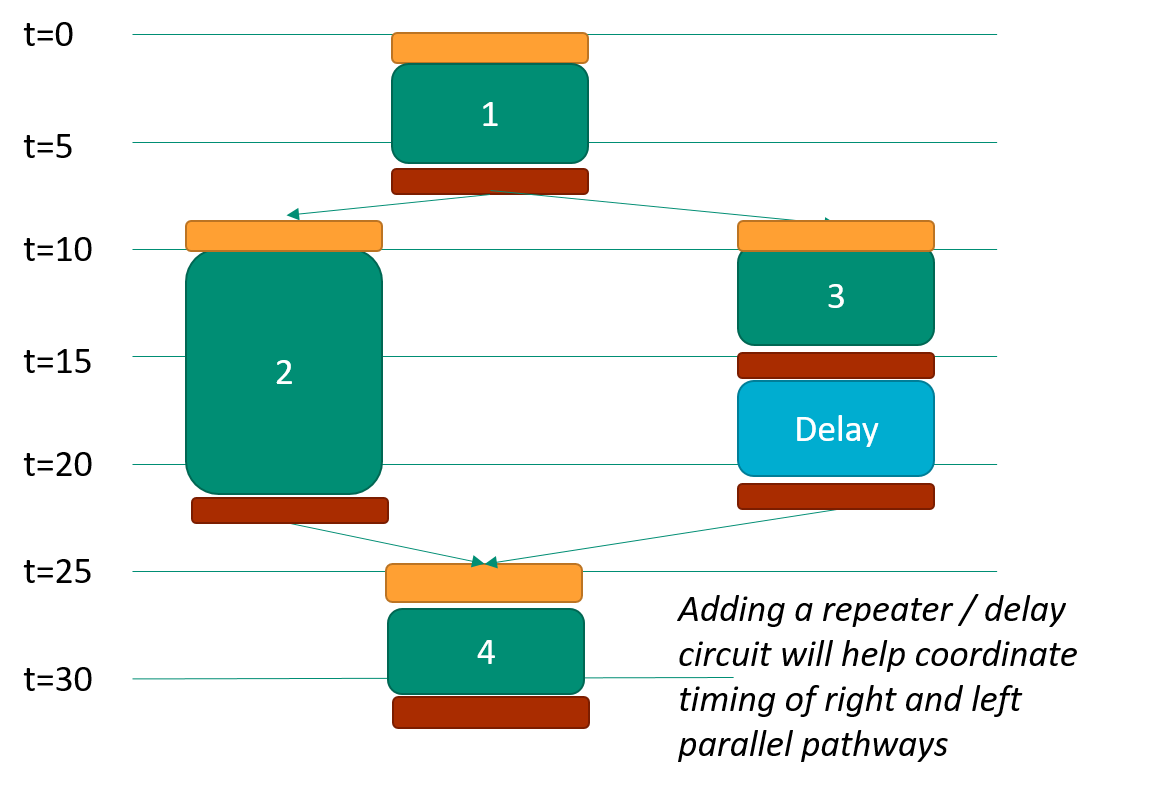}
	\caption{Adding a delay to synchronize Fugu branches}
	\label{parallel_link}
\end{figure}

\paragraph{Align timings of bricks} Each of the bricks requires some amount of computation time or `circuit depth'.  If bricks are intended to run in \textit{parallel}, the difference in timings and depths may require that an additional \textit{delay circuit} is instantiated on one of the pathways to ensure that the circuit is appropriately timed. Figure \ref{parallel_link} illustrates this idea. 

As the branches may be indexed arbitrarily and the time or depth of a module may be undetermined until its overall size is identified, it is unknown at the start of Fugu which branch will be the limiting factor in terms of time.  This longest branch is the reference length that all other branches must match.  Once this \textit{branch depth} is found, we then work through each of the other branches to make the depths equivalent.

If the branch depth can be fully determined at build time, we can simply add a delay block - most simply a set of repeater neurons that spike with a delay of whatever the difference is.  Most simply, this delay block could be at the end of the branch. However, there is likely a benefit to load balance over time; the delays will be relatively cheap in terms of computation, and thus they can perhaps be staggered at different times of each branch to keep the overall network activity at a roughly comparable level.  

If the branch depth is variable or dependent on the input data (e.g. an iterative algorithm with depth determined at runtime), then we can stage information other branches in a buffer until all branches have completed their computation.  This buffer can then be released using control nodes---Extra neurons instantiated in the graph to signal control commands.  Each brick defines at least one control node that fires on completion.  This signal can then be used to flush a buffer and continue the computation.

There is also a third potential solution, though it would require more careful implementation.  Many of the SNAs being developed can be tailored to use fewer neurons if more time is available.  This time-space tradeoff is generally biased towards the faster algorithm; however in cases where a module with such a tradeoff sits within a branch with ``free-time'' so to speak, it is possible, and perhaps even advantageous, to represent that module in a more time-costly, space-efficient manner that reduces the overall neuron footprint of the model.

\begin{figure*}
	\includegraphics[width=5in]{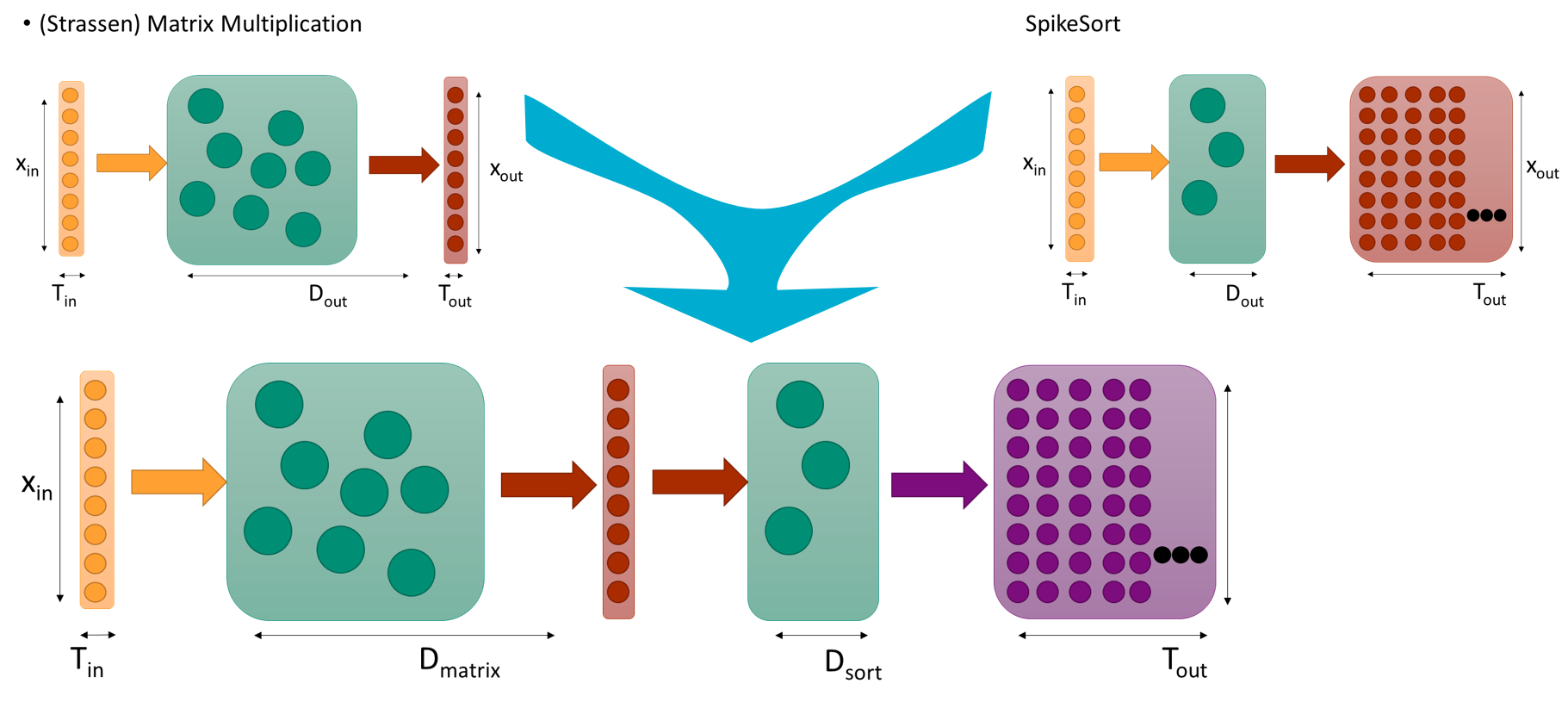}	
	\caption{Combined Bricks}
	\label{combined}
\end{figure*}

\subsubsection{Output}
The output of Fugu will be a single NetworkX graph that fully describes the spiking neural algorithm.  The \textbf{edges} of this graph will be the synapses of the model, and will accordingly have weights associated with them as attributes.  The \textbf{nodes} of this graph will be the neurons of the model, and will accordingly have dynamics parameters associated with them.  Additionally, some parameters, such as learning rates, additional dynamical states, etc. may be included within the attributes.

\subsection{Neuromorphic Hardware Compilers, the Reference Simulator, and Intentional Limitations of Fugu}
The primary output of Fugu is the above-stated NetworkX graph that represents the desired neural algorithm, but is also hardware agnostic.  Currently, each major neuromorphic platform has its own unique programming interface, and thus it is reasonable to assume that for the near future the need for a ``neuromorphic hardware compiler'' will be met by one-off solutions specific to different platforms.

Because of this, we envision that as these hardware-specific interfaces begin to stabilize, it will hopefully be straightforward to convert from this generic NetworkX computational graph description of the Fugu algorithm to any required hardware format.  However, given that these platforms remain a fluid target, Fugu also includes a \textit{reference simulator} which enables algorithms to be prototyped. 

The reference simulator is designed to be conservative: clearly the breadth of what simulations can implement far outstrip current neuromorphic hardware capabilities today; especially when dynamics such as synaptic plasticity and architectural reconfiguration are considered.  As such, because the Fugu simulator is intended to show that an implemented algorithm would be capable of running on a generic neuromorphic system, we do not yet include learning, complex neuron types, or other capabilities beyond the basic LIF model.  This is not a judgment on the value or importance of those models; but rather a reflection of the state of the hardware community.  As these platforms capabilities universally begin to move to include these dynamics, the reference simulator will advance accordingly to encompass these capabilities. Furthermore, the reference simulator is not intended to provide indications of the runtimes and other performance measures neuromorphic hardware can enable.   

Meanwhile, Fugu bricks can be constructed that include node (neuron) and edge (synapse) attributes that relate to learning rules or other extended dynamics, but these would not be assured to operate as expected on all neuromorphic platforms.

\section{Examples}
As follows are four illustrative examples of how bricks and scaffolds may be defined in Fugu to compute a suite of computations ranging from basic logic operations to higher level applications such as graph analytics or scientific computing. In these examples, different principles of the spiking dynamics are utilized to infer the result of the computation. This may include the times when spikes occur as well as the specific neurons which spike. 

\subsection{Logic}
As a simple example brick, we first present the logical AND function. This canonical logic operation outputs true (or one) only when all of its inputs are simultaneously true. A neuron is able to compute this operation when nominal inputs each contribute to the receiving neuron whose threshold is equal to the number of inputs. A leakage value causing the neuron to return to a resting potential of zero every time-step resets the neuron requiring all inputs must be received simultaneously. While simplistic, this intuitive brick example illustrates the definition of neuron parameters and can be composed in more elaborate scaffolds capable of executing Boolean logic functions. 

Additionally, this brick is easily modified to perform a logical OR.  You simply lower the threshold value so that any incoming spike is enough to have the OR neuron fire.  In this case, as with the AND neuron, the boolean function is computed exactly and can be used across coding schemes as long as the shapes of the incoming bricks are equal.

In general, we do not expect efficiency gains (either wall-time or theoretical) from implementing boolean logic on neuromorphic systems.  However, these basic components (along with other simple functions) are key to creating larger, more sophisticated data flows all while staying on-hardware, avoiding costly I/O with a host system.

\subsection{Graph Analytics}
Given a graph, a core computation is determining the distance between two nodes. We can easily instantiate this as a Fugu brick by taking a target graph (i.e., the graph on which we are determining distances) and creating a spiking neural network where each node is represented by a neuron and each graph edge is a synapse.  For a un-directed graph, we create a synapse in both directions.  For a weighted graph, the weight is converted to a proportional synaptic delay.  Then, we simply set the parameters such that (1) a neuron fires when receiving any spike and (2) a neuron only fires once (this can be done by a very low reset voltage or by a self-link with strong negative weight).  Then, we begin the algorithm forcing the source node/neuron to spike.  The shortest path distance between the source neuron and any other neuron is computed by the time it takes for the target neuron to spike.  This algorithm has been described independently several times, including~\cite{hamilton2019spike}, and our inclusion here is not novel.  However, we do now remark how such an algorithm could fit within a larger Fugu scaffold for more complicated algorithms.  

First, we note that the algorithm is `started' by having the source node fire.  Fugu can provide an input with equivalent dimensionality as the nodes on the graph, essentially a rasterized image of the target graph in neurons.  This means that we can use Fugu-style preproccesing functions to determine the source neuron.   A simple example of such would be using an image classifier to identify points of interest in overhead imagery.  The user can then cross-reference that overhead imagery with a map of the area, and Fugu effectively computes distances on that map according to the location of objects in the imagery.

Second, the output of the path-length determination is temporally coded---The amount of time it takes for the target neuron to spike is equal to the length of the shortest path from the source node to the target node.  This temporal representation is one of the core number schemes supported in Fugu.  As such, we can then take the output of this brick and perform in-Fugu post-processing.  For example, we can easily compare the value against a reference value using a Threshold brick (e.g., computing if one has enough fuel to make a trip) or compare against other path lengths in a first-come-first-serve Minimum brick (e.g., determining who would get to a destination first).

These handful of example extensions only being to describe the types of complex algorithms possible through compositionality.  Despite each of the above bricks being simple in there own right, they can combine together for sophisticated calculations.

\subsection{Game Theory}
Building upon the prior two examples, cast to a temporal regime, a neural computation graph can be used to determine the pure strategy Nash equilibra of some game theoretic problems. For example, consider the canonical two player prisoner's dilemma game. In this scenario, two captured prisoners are interrogated separately and the prison sentences they receive are dictated by their decisions to confess and testify against their partner or to not confess and stand firm in their allegiance to their criminal partner. Each criminal has preferences over their sentencing, namely to receive a shorter sentence, and the game dynamics are defined such that what is best for an individual prisoner is not the same as what is best for the two criminals jointly. 

Representing all pairs of action selections as neurons, and the player preferences as the distances between a player neuron and the action space neurons, a graph for determining pure strategy Nash equilibria may be built. The Nash equilibria is a solution concept identifying outcomes in which no single player can improve upon the result by changing their own strategy. To identify these equilibrium points, an action neuron can only fire if it receives a spike from both players. This may be achieved either by constructing a scaffold using AND bricks such that both inputs must be received simultaneously, or by using the graph analytic distance between node dynamics to incorporate a firing threshold of two and full leakage every timestep. The later approach essentially embeds a logical AND function as the neuron activation function with the connectivity structure of the graph imposing that the inputs come from separate input source nodes. 

In this sense, while individual players may prefer other outcomes and effectively send spikes to their closer preferred actions sooner in time, it is only the joint activation of player actions which drive a neuron to fire. And effectively, the determination of the Nash equilbrium is computed by the spiking of action pair neurons. This premise may be extended to include more actions as well as larger action spaces, however for clarity of explanation we described the simple two player two action game here.  

\subsection{Scientific Computing}
Our final example is an illustration of how a standalone neural algorithm, in this case targeting scientific computing, can be broken down into separate bricks that potential could have alternative uses in contributing to other algorithms.  Previously, we described two neural algorithms for simulating a Markov random walk process to approximate the diffusion equation \cite{severa2018spiking}.  One of the algorithms, termed the \textit{particle method} in our previous study, simulated random walk particles directly by dedicating a sub-population of neurons to simulate each walker.  A solution to the diffusion equation (and in turn many other partial differential equations for which diffusion is a key kernel) can then be achieved by taking population statistics of the random walkers at different time-steps.

The random walk particle model conceived in \cite{severa2018spiking} is a monolithic neural circuit consisting of many walkers that each consist of several rings of neurons and control neurons that implement the random movements.  In effect, this model combines two functions that are repeated for each walker - random movement neurons and a population that tracks position.  

\begin{figure}
\includegraphics[width=2.8in]{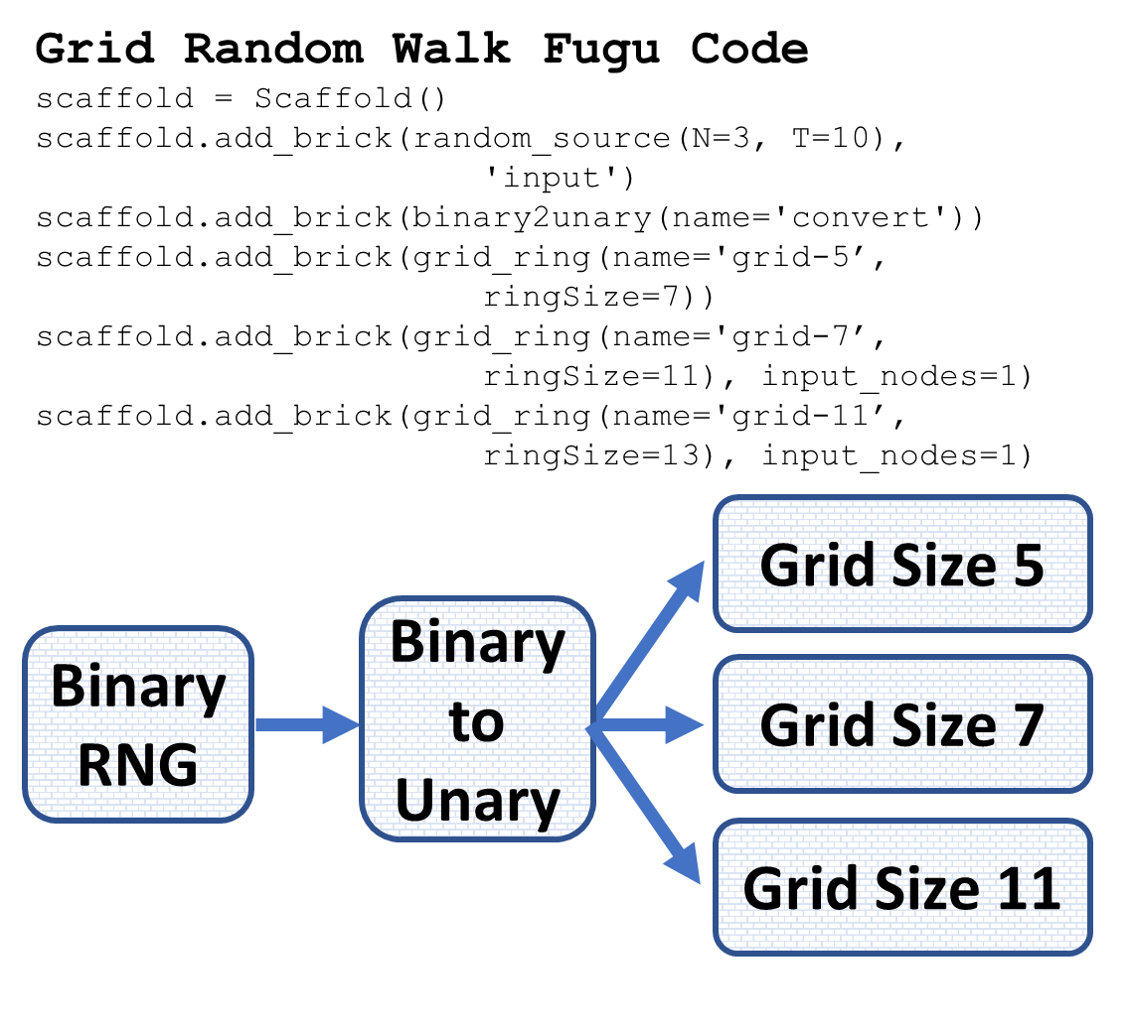}
\caption{Schematic of Grid Random Walk Implementation in Fugu.  Each block corresponds to a distinct circuit generated by Fugu which are then composed together to simulate the random walk process.}
\label{GridDiagram}
\end{figure}

The Fugu implementation, as summarized in Figure \ref{GridDiagram} breaks this monolithic neural circuit into three separate components.  For simplicity, what is illustrated is a \textit{grid} torus that tracks position over a hexagonal grid, inspired by the grid cells in the brain and the recent finding that biological navigation leverages random walks as well \cite{sreenivasan2011grid, stella2019hippocampal}.  In Fugu, we separate the random walkers into three parts: a \texttt{binary random number generator} brick, a \texttt{binary to unary} conversion brick, and a \texttt{grid position tracker} brick, the latter of which will be repeated at several different sizes for each random walker that is moving over two dimensions.  

Notably, by breaking the random walk algorithm into these components, future design decisions rapidly become evident.  Depending on the application needs and algorithm / hardware efficiency, a final brick could be designed that implements the Chinese Remainder Theorem to convert from the grid modular code to a histogram of random walker locations over space.  Alternatively, a hippocampus-inspired place cell brick could be designed that perform this conversion approximately.

\section{Discussion $\&$ Conclusions}

The Fugu framework, as introduced here, provides a powerful ability to abstract the underlying neural dynamics which emerging neuromorphic hardware is striving to enable. Not only does this abstraction and composition capability enable a means by which non-experts can consider the benefits of SNAs in their computations, but it also helps to enable the exploration of when it is advantageous to do so. In some cases, it has been shown that a SNA approach is only theoretically beneficial for problems of specific scale. Fugu enables the composition of SNAs such that individually their benefits may be limited, but the resultant composite computation may be favorable factoring in full circuit costs. For example, Figure \ref{combined} shows the combination of bricks for Strassen matrix multiplication and SpikeSort which together may yield the computation of a sorted multiplication. 

Leveraging the analysis of algorithm composition Fugu enables, it may also help provide insight into the impact of architecture specific design choices when mapping to specific hardware platforms. For example, if architecture limitations necessitate extensive use of repeater or delay circuitry that exceeds either the available resources of the architecture or drive up the runtime of the computation these insights may suggest another hardware be considered or alternative algorithmic formulation to meet performance considerations. Alternatively, it is also possible that by leveraging the Fugu provided IR, neuromorphic compilers may explore additional optimization of the circuit layout which the specific architectural features of a given neuromorphic platform may be able to uniquely enable. 

By providing Fugu as a framework for developing and composing spiking algorithms in terms of computation graphs we strive to further the development of SNAs, enable non-experts to utilize neuromorphic computing in a variety of domains such as scientific computing, and iterate with neuromorphic architecture development to jointly understand the impact of design choices both algorithmically and architecturally. 

\section*{Acknowledgment}
This research was funded by the Laboratory Directed Research and Development program at Sandia National Laboratories and the DOE Advanced Simulation and Computing program.  Sandia National Laboratories is a multi-mission laboratory managed and operated by National Technology and Engineering Solutions of Sandia, LLC., a wholly owned subsidiary of Honeywell International, Inc., for the U.S.~Department of Energy's National Nuclear Security Administration under contract DE-NA0003525.

This article describes objective technical results and analysis.  Any subjective views or opinions that might be expressed in the paper do not necessarily reflect the views of the US Department of Energy or the US Government.

\bibliographystyle{ACM-Reference-Format}
\bibliography{ICONS}

\end{document}